\documentclass[runningheads]{llncs}
\usepackage[T1]{fontenc}
\usepackage{graphicx}
\usepackage{marvosym}
\begin{document}
\title{Unsupervised MR-US Multimodal Image Registration with Multilevel Correlation Pyramidal Optimization}
%
%
\author{Jiazheng Wang\inst{1, 2} \and
Zeyu Liu\inst{1, 2} \and
Min Liu\inst{1, 2}\textsuperscript{(\Letter)} \and
Xiang Chen\inst{1, 2} \and 
Xinyao Yu\inst{3} \and 
Yaonan Wang\inst{1, 2} \and
Hang Zhang\inst{4}}
\authorrunning{J. Wang et al.}
%
\institute{School of Artificial Intelligence and Robotics, Hunan University, Changsha, Hunan, China \and
National Engineering Research Center of Robot Visual Perception and Control Technology, Hunan University, Changsha, Hunan, China \\
\email{\{wjiazheng, liuzeyulzy, liu\_min, xiangc, yaonan\}@hnu.edu.cn}
\and National University of Singapore, Singapore\\
\email{xinyao.yu@u.nus.edu}
\and Cornell University, USA \\
\email{hz459@cornell.edu}}
\maketitle              
\begin{abstract}

Surgical navigation based on multimodal image registration has played a significant role in providing intraoperative guidance to surgeons by showing the relative position of the target area to critical anatomical structures during surgery. However, due to the differences between multimodal images and intraoperative image deformation caused by tissue displacement and removal during the surgery, effective registration of preoperative and intraoperative multimodal images faces significant challenges. To address the multimodal image registration challenges in Learn2Reg 2025, an unsupervised multimodal medical image registration method based on 
Multilevel Correlation Pyramidal Optimization (MCPO) is designed to solve these problems. First, the features of each modality are extracted based on the modality independent neighborhood descriptor, and the multimodal images is mapped to the feature space. Second, a multilevel pyramidal fusion optimization mechanism is designed to achieve global optimization and local detail complementation of the displacement field through dense correlation analysis and weight-balanced coupled convex optimization for input features at different scales. Our method focuses on the ReMIND2Reg task in Learn2Reg 2025. Based on the results, our method achieved the first place in the validation phase and test phase of ReMIND2Reg. The MCPO is also validated on the Resect dataset, achieving an average TRE of 1.798 mm. This demonstrates the broad applicability of our method in preoperative-to-intraoperative image registration. The code is available at https://github.com/wjiazheng/MCPO.

\end{abstract}
\section{Introduction}

Medical image registration has been an important topic in the field of medical image analysis, and many significant methods \cite{ref_article4,ref_article5,ref_article6,ref_article8} have driven the development of medical image registration tasks. Deep learning-based medical image registration methods \cite{ref_article7} generally involve long and complex learning processes, and often struggle to achieve accurate estimation for multimodal, large-deformation data and general usability for extensive tasks. The Learn2Reg 2025 sub-challenge, ReMIND2Reg \cite{remind}, is a multimodal medical image registration task oriented to preoperative Magnetic Resonance Imaging (MRI) and intraoperative ultrasound (US), which is characterized as unlabeled, large deformation, and low feature distinctness. Aiming at the above characteristics, inspired by \cite{ref_proc1,ref_article1}, an unsupervised multimodal medical image registration method based on Multilevel Correlation Pyramidal Optimization (MCPO) has been proposed, which can quickly achieve effective MR-US multimodal image registration using only a small number of learning and optimization procedures.

\section{Methodology}

The proposed MCPO method is based on VoxelOpt \cite{voxelopt} and ConvexAdam \cite{ref_url1} with a series of improvements, which include (1) Introducing a weight-balancing term on coupled convex optimization to achieve smoother deformation optimization. (2) A multilevel pyramidal fusion optimization mechanism is designed to achieve coarse-to-fine representation of the dense displacement field by performing affine transformation and fusing convex optimization results across different scales. (3) Finally, fine-tuning of the rigid displacement field is achieved through an optional Adam instance optimization.

\begin{figure}
\includegraphics[width=\textwidth]{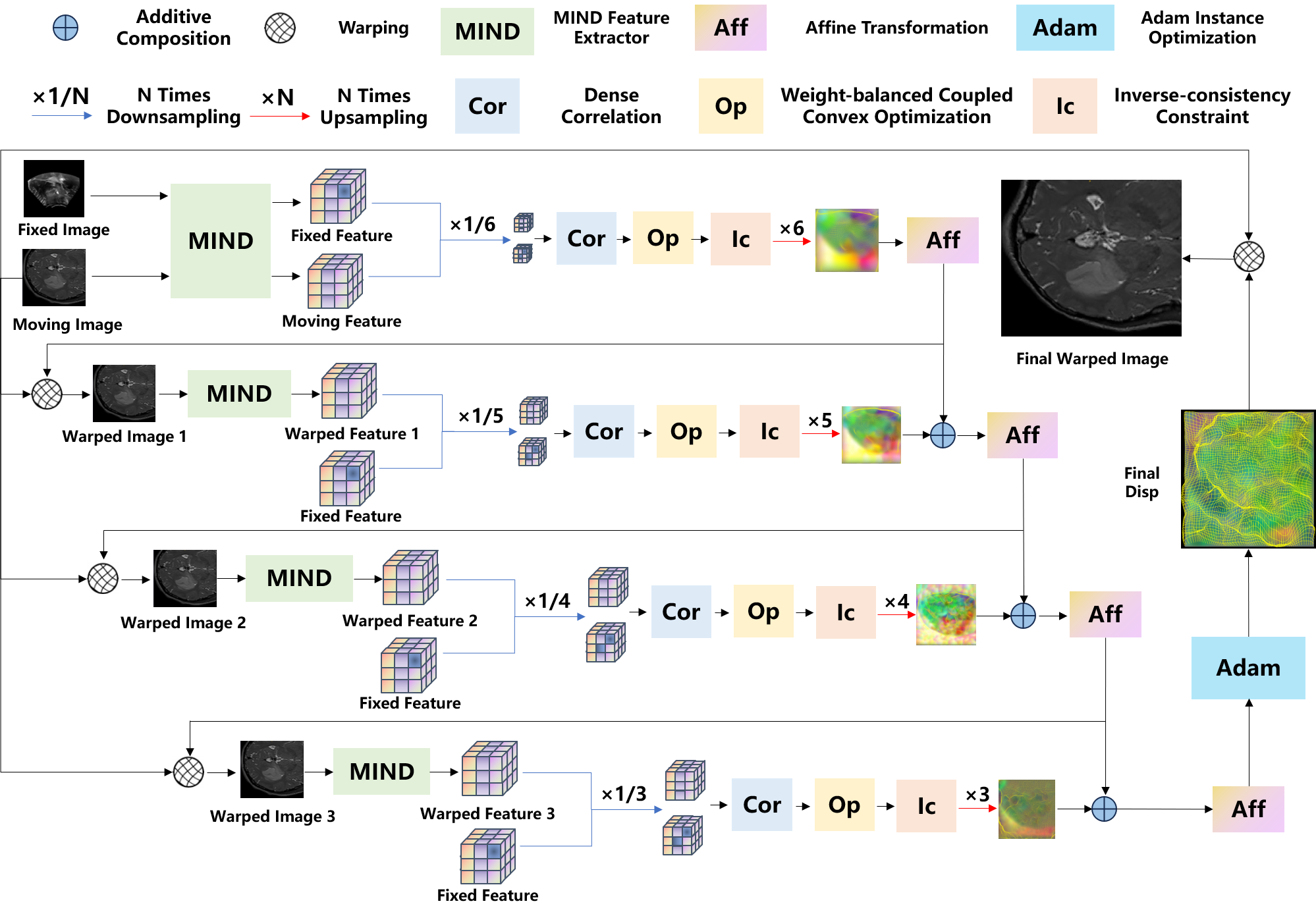}
\caption{The overall flow of the proposed MCPO method.} \label{fig1}
\end{figure}

The overall flow of the proposed MCPO method is shown in Fig.~\ref{fig1}. The moving image and the fixed image are inputted and then the modal-independent features of the images are first obtained by Mind-SSC Feature Extractor \cite{ref_article1}. The Mind-SSC feature extractor exploits the self-similarity of the partial area in the image to extract the unique structural information of the local neighborhood, which results in a highly consistent structural representation across modalities.

Inspired by VoxelOpt \cite{voxelopt}, the acquired features are subsequently downsampled by average pooling operations at different times $n$ ($n=3,4,5,6$) to obtain features in different scales $F_{n}$ as inputs for the multilevel pyramidal fusion optimization mechanism. For each level, the acquired input features $F_{n}$ are first fed into the dense correlation layer to compute the sum-of-squared-differences (SSD) cost volume and the initial optimal displacements for each voxel. The large search space allows us to make an initial capture of the displacement for each voxel, even though some voxel points may have large deformations. The output of the dense correlation layer is alternately optimized for similarity and smoothness by iterations in the weight-balanced coupled convex optimization layer, and then the iterative optimization results are averaged to achieve global regularization with weight balancing. After that, an inverse consistency constraint layer \cite{ref_proc2} is introduced to minimize the difference between the forward and backward transformations to avoid unrealistic deformations, obtaining the initial displacement field of each level. Subsequently, an affine transformation matrix is computed via least-squares fitting to the initial displacement field, yielding a rigid displacement field at each level. The dense displacement field is constructed in a coarse-to-fine manner through additive composition of the rigid displacement field with the subsequent-level field. After completing the additive composition of displacement fields at each level, the final rigid displacement field is obtained. 

Furthermore, an optional Adam instance optimization \cite{ref_proc1} is applied to the final rigid displacement field to further refine the field. In addition to the regularization loss and the Mind loss used in the original Adam instance optimization, we also introduce a stochastic patch-based mutual information loss. Due to the low information density in ultrasound images, the traditional mutual information loss, which relies on global image computations, fails to perform effectively in this task. Therefore, the stochastic patch-based mutual information loss refines the displacement field by randomly sampling informative local patches and comparing their mutual information, thereby leveraging similarities among effective features. Depending on whether Adam instance optimization is employed, two variants of the method are proposed: \textbf{MCPO-rigid} and \textbf{MCPO-deform}. Their performance is experimentally evaluated and analyzed in the subsequent section.

\section{Experiments and Results}

Our method focuses on the ReMIND2Reg sub-challenge in Learn2Reg 2025, and to verify the generality of the method, we also tested it on the Resect dataset.

\subsection{Dataset}

\paragraph{ReMIND2Reg Sub-challenge.}
The dataset of ReMIND2Reg sub-challenge is a pre-processed subset of the ReMIND dataset \cite{ref_article2}, which contains pre- and intra-operative data collected on consecutive patients who were surgically treated with image-guided tumor resection between 2018 and 2024. The goal of the ReMIND2Reg sub-challenge is to register pre-operative MRI from multiple modalities (including ceT1 and T2) and intra-operative 3D ultrasound images. Common pre-processing to the same voxel resolution ($0.5\times0.5\times0.5$ mm) and spatial dimension ($256\times256\times256$) is performed on all the images. The training set has 99 patients with 99 3D ultrasound images, 93 ceT1 images and 62 T2 images, and the validation set contains 5 patients with 5 images for each modality. The test set is unseen for the participants. 

\paragraph{Resect Dataset.}
The Resect dataset \cite{resect} comprises intraoperative ultrasound images and preoperative MRI images from 23 patients with low-grade gliomas. It includes 23 ceT1 MRI scans and 23 FLAIR MRI scans as the moving images. Annotation data for validation are also provided, with each image having dimensions of $256\times256\times288$ voxels. The patient cohort was selected without significant bias, encompassing tumors located in diverse brain regions, which ensures commendable representativeness and diversity. This dataset serves as an effective supplementary resource for validating the performance of different methods.

\subsection{Implementation Details}

For MCPO-rigid, the radius and dilation of the Mind-SSC feature extractor are fixed to 1 and 2, respectively, and the displacement range of the discretised search space is set to 4. For MCPO-deform, the radius and dilation of the Mind-SSC feature extractor are fixed to 3 and 3, respectively, and the displacement range of the discretised search space is set to 6. For the Adam instance optimization in MCPO-deform, the number of iterations is set to 20, the smooth convolution kernel is set to 5, the weight of the stochastic patch-based mutual information loss is set to 100, and the rest of the parameters are referred to the original settings \cite{ref_url1}. These are the parameters that deliver optimal performance with limited runtime determined through multiple experiments. 

We employed ConvexAdam \cite{ref_url1} with affine transformation, NiftyReg \cite{ref_url2}, and MCBO \cite{mcbo} as the baseline methods.
The evaluation of registration performance in our paper is based on Target Registration Error (TRE), which is defined as the Euclidean distance between corresponding landmarks in the fixed image and the warped moving image. All experiments were conducted using PyTorch, with the instance optimization performed on a single NVIDIA RTX 3090 GPU.

\subsection{Experimental Results}

\begin{figure}
\includegraphics[width=\textwidth]{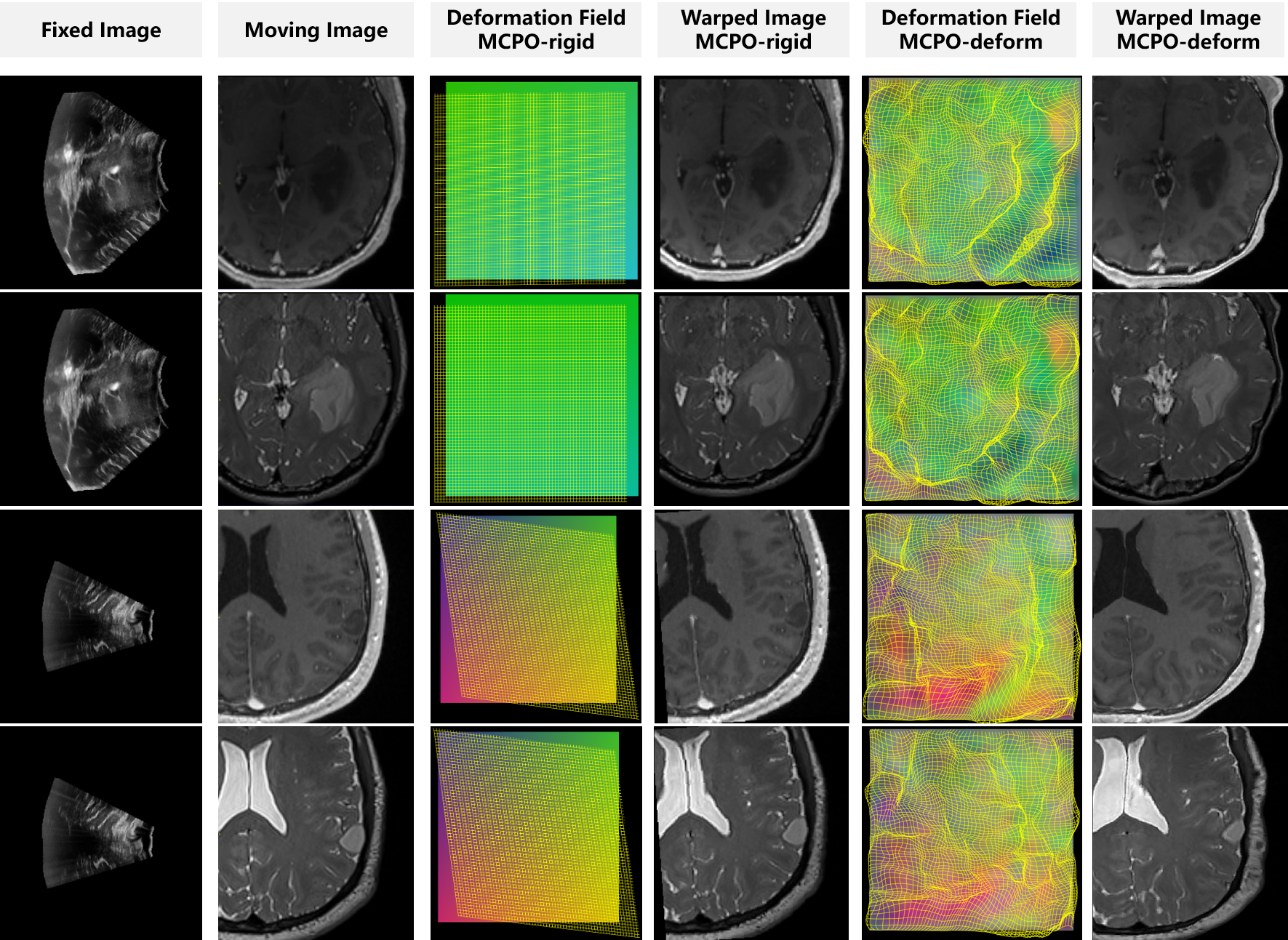}
\caption{Visualization results of ReMIND2Reg sub-challenge in Learn2Reg 2025.} \label{fig2}
\end{figure}

\paragraph{Validation Phase of ReMIND2Reg Sub-challenge.}
The experimental results are shown in Table~\ref{tab1}. The visualization of the multimodal image registration for this task is shown in Fig.~\ref{fig2}. On the validation set of the ReMIND2Reg Sub-challenge, MCPO-rigid demonstrated optimal performance, achieving a TRE of 1.790 $\pm$ 0.536 mm. Although the overall result of MCPO-deform was moderate due to suboptimal performance on a specific case, it remained comparable to baseline methods and marginally outperformed NiftyReg.

\begin{table}
\centering
\caption{Results of ReMIND2Reg sub-challenge in Learn2Reg 2025.}\label{tab1}
\begin{tabular}{c|c}
\hline
Methods  &  TRE(mm) \\
\hline
Initial &  3.746 $\pm$ 0.639\\
ConvexAdam-Rigid &  2.609 $\pm$ 1.217\\
NiftyReg & 2.807 $\pm$ 1.228\\
MCBO & 2.367 $\pm$ 0.638\\
MCPO-rigid & \textbf{1.790 $\pm$ 0.536} \\
MCPO-deform & 2.766 $\pm$ 1.001 \\
\hline
\end{tabular}
\end{table}

\begin{figure}
\includegraphics[width=\textwidth]{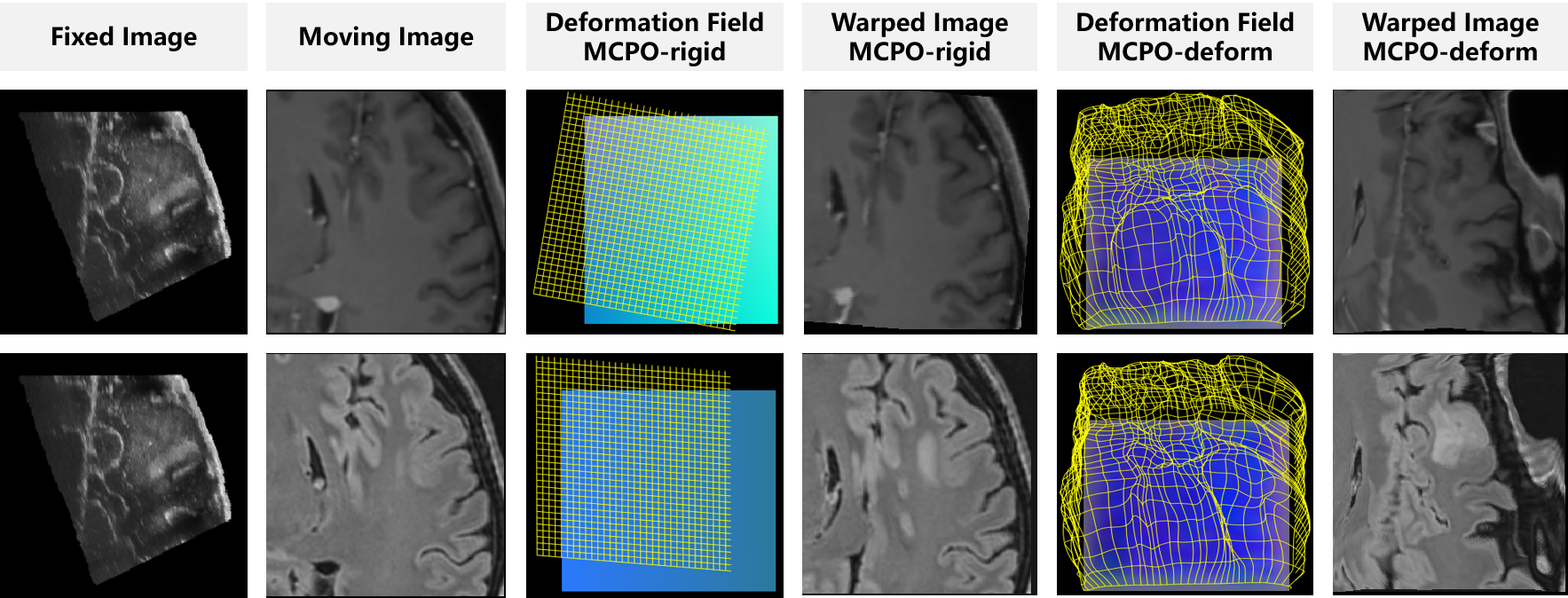}
\caption{Visualization results of a large deformation case in the Resect dataset using MCPO-deform.} \label{fig3}
\end{figure}

\begin{figure}
\includegraphics[width=\textwidth]{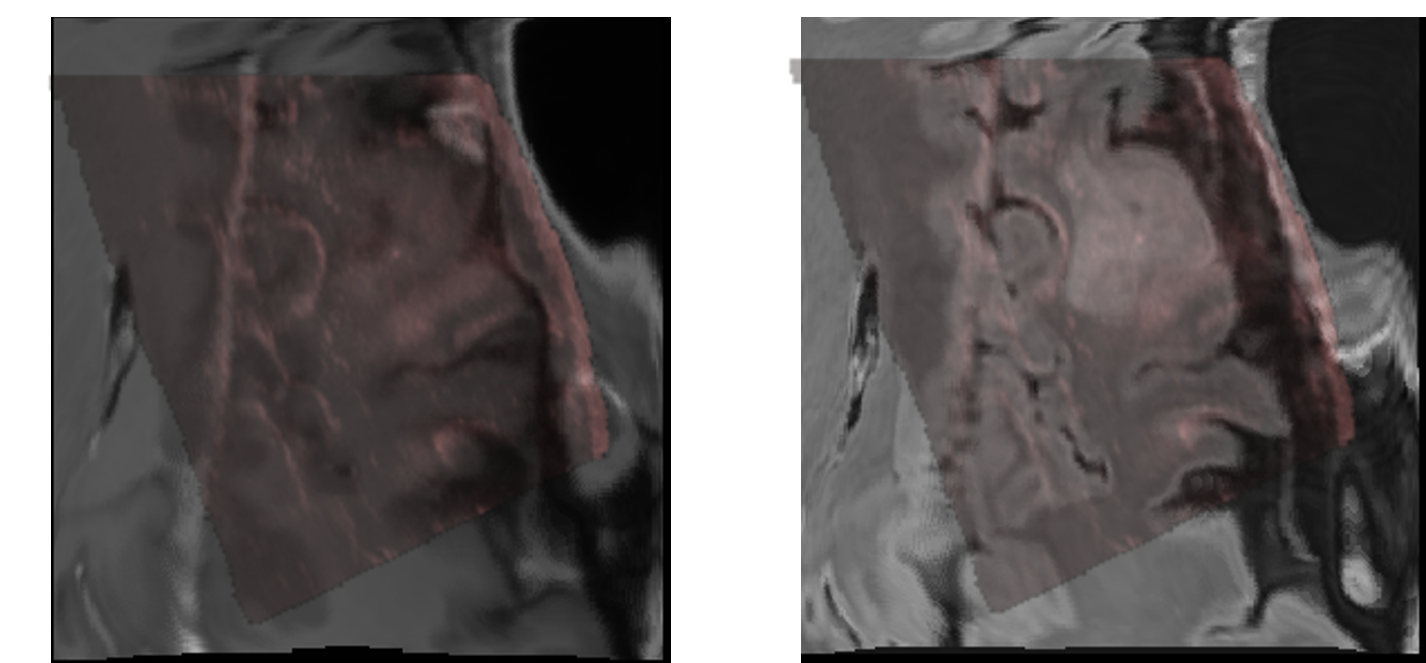}
\caption{Visualization result of MRI-US fusion for the large deformation case before (left) and after (right) registration using MCPO-deform.} \label{figl}
\end{figure}

\paragraph{Resect dataset.}
Since the ReMIND2Reg Sub-challenge includes only five patients (ten cases) with annotation information as the validation set with all exhibiting relatively small deformations, validation based solely on this dataset may lead to potential overfitting to a limited number of similar data, making it difficult to comprehensively evaluate the performance of the methods. Therefore, we conducted additional experiments on the Resect dataset, with results presented in Table~\ref{tab2}. On this more diverse dataset, MCPO-deform achieved superior average TRE of 1.798 $\pm$ 1.301 mm, while other methods often failed to handle cases with large deformations effectively. As illustrated in Fig.~\ref{fig3}, which shows a case from the Resect dataset with an initial deformation of 19.731 mm, MCPO-deform successfully aligned key anatomical regions between the fixed and warped images. For this case, MCPO-deform achieved registration errors of 1.136 mm (FLAIR to ultrasound) and 1.368 mm (ceT1 to ultrasound). In comparison, MCPO-rigid only reached 13.313 mm and 16.195 mm, respectively. The visualization result for this case is shown in Fig. \ref{figl}. Considering these results, we ultimately submitted MCPO-deform as the Docker for the testing phase of the ReMIND2Reg sub-challenge.

\begin{table}
\centering
\caption{Results of methods in Resect dataset.}\label{tab2}
\begin{tabular}{c|c}
\hline
Methods  &  TRE(mm) \\
\hline
Initial &  5.374 $\pm$ 4.173\\
ConvexAdam-Rigid &  2.673 $\pm$ 1.474\\
NiftyReg & 2.635 $\pm$ 3.009\\
MCBO & 3.785 $\pm$ 3.716\\
MCPO-rigid & 2.941 $\pm$ 2.939 \\
MCPO-deform & \textbf{1.798 $\pm$ 1.301} \\
\hline
\end{tabular}
\end{table}

\paragraph{Test Phase of ReMIND2Reg Sub-challenge.}

Upon submission the Docker to the test phase, the organizer uniformly tested all participating teams’ results, as illustrated in the figure. The competition officially introduced an additional metric, TRE30, defined as the 30th percentile of the TRE values computed across all landmarks, which further quantifies the registration performance after excluding potential outlier cases.

\begin{figure}
\includegraphics[width=\textwidth]{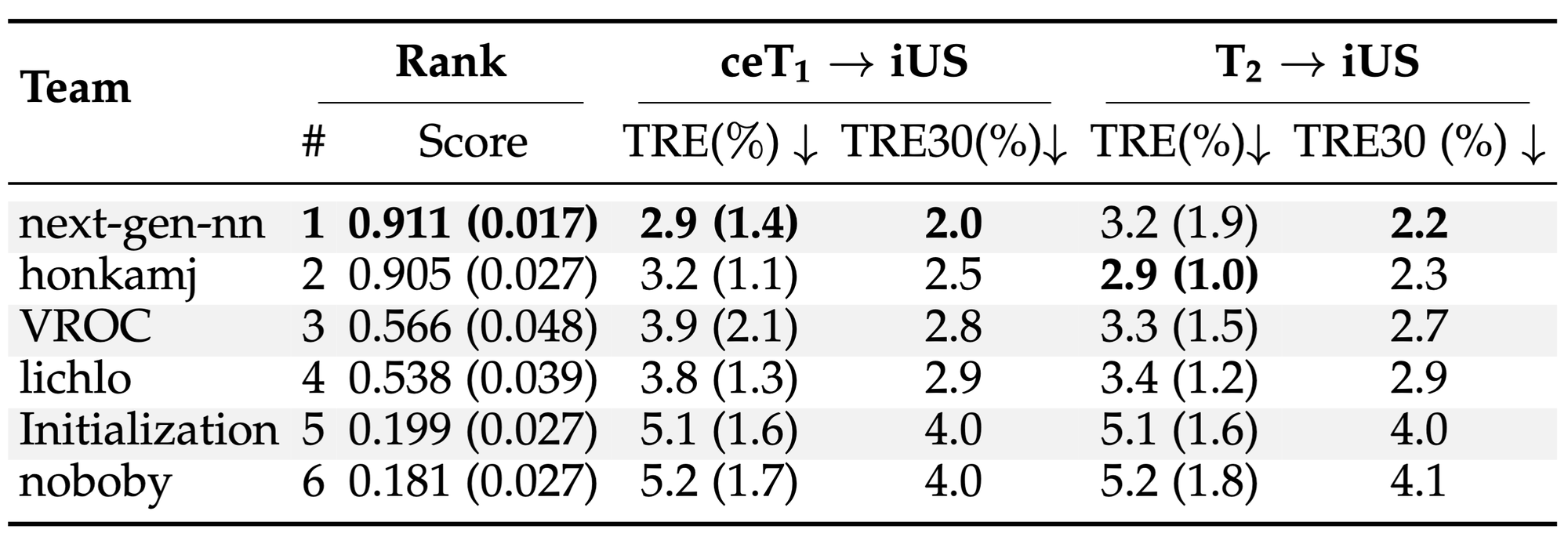}
\caption{Results of methods in the test phase of ReMIND2Reg Sub-challenge. Reported by the organizer.} \label{fig25}
\end{figure}

As shown in the Fig. \ref{fig25}\footnote{https://learn2reg.grand-challenge.org/}, the MCPO-deform, corresponding to the results of team Next-gen-nn, achieves superior performance across most evaluation metrics. It should also be noted that the method from the second-placed team also demonstrates highly competitive registration accuracy. After normalizing all results, the proposed method attained a final score of 0.911 (out of 1.0), securing first place in the ReMIND2Reg Sub-challenge by a narrow margin. This outcome further validates that the proposed method exhibits enhanced generalization capability for preoperative‑to-intraoperative registration and can effectively adapt to diverse and complex clinical scenarios.

\section{Conclusion}
The application of the proposed MCPO method to the Learn2Reg 2025 challenge shows that a multilevel optimization strategy using only a small amount of learning can quickly and accurately achieve the registration between multimodal medical images with large deformations. Meanwhile, the method proposed in this paper achieves promising results in Resect dataset, which illustrates the generality of the method for multimodal medical image registration.

\begin{credits}

\subsubsection{\ackname} Thanks all the organizers of the MICCAI 2025 Learn2Reg challenge. This work was supported by the National Natural Science Foundation of China under Grant U22B2050, 62425305, and 62221002. 

\end{credits}

%
%
%
%

\end{document}